\documentclass[12pt]{article}
\usepackage{setspace}
\usepackage{enumitem}

\usepackage{graphicx,color}
\usepackage{subcaption}

\usepackage{lscape}
\usepackage{wrapfig}
\usepackage{tikz}
\usepackage{float}
\usepackage{multirow}
\usepackage{amsmath, amssymb, amsfonts, amsthm}
\usepackage{bm}
\usepackage{dsfont}
\DeclareMathOperator{\Tr}{tr} 
\DeclareMathOperator{\sign}{sign} 
\renewcommand{\vec}[1]{\mathbf{#1}} 
\newcommand{\mat}[1]{\mathbf{#1}} 
\newcommand{\trans}[1]{#1^\mathsf{T}} 
\newcommand{\st}{\quad \mathrm{s.t.}\ } 
\newcommand{\rank}[1]{\mathrm{rank}\left(#1\right)} 
\providecommand{\norm}[1]{\left\lVert#1\right\rVert} 

\newcommand{\St}[2]{\mathrm{St} \left(#1, #2\right)}
\newcommand{\StG}[2]{\mathrm{GSt} \left(#1, #2\right)}

\numberwithin{equation}{section}

\usepackage[noend]{algpseudocode}
\usepackage{algorithm}


\makeatletter
\newcommand{\subsubsubsection}{\@startsection{paragraph}{4}{\z@}%
{1.5\baselineskip \@plus.5\dp0 \@minus.2\dp0}%
{.5\baselineskip \@plus2.3\dp0}%
{\reset@font\normalsize\bfseries}
}
\newcommand{\subsubsubsubsection}{\@startsection{subparagraph}{5}{\z@}%
{1.5\baselineskip \@plus.5\dp0 \@minus.2\dp0}%
{.5\baselineskip \@plus2.3\dp0}%
{\reset@font\normalsize\itshape}
}
\makeatother
\usepackage[subrefformat=parens]{subcaption}
\usepackage[subrefformat=parens]{caption}

\makeatletter
\def\eqnarray{\stepcounter {equation}\let \@currentlabel =\theequation
\global \@eqnswtrue
\global \@eqcnt \z@ \tabskip \@centering \let \\=\@eqncr
$$\halign to \displaywidth \bgroup \@eqnsel \hskip \@centering
$\displaystyle \tabskip \z@ {##}$&\global \@eqcnt \@ne \hfil
${\mbox{}##\mbox{}}$\hfil &\global \@eqcnt \tw@
$\displaystyle \tabskip \z@ {##}$\hfil \tabskip \@centering
&\llap {##}\tabskip \z@ \cr}
\makeatother

\usepackage{url}

\begin{document}

{\baselineskip = 8mm 

\begin{center}
\textbf{\LARGE Sparse Reduced-Rank Regression for Simultaneous Rank and Variable Selection via Manifold Optimization} 
\end{center}

\begin{center}
{\large Kohei Yoshikawa$^{1}$, \ Shuichi Kawano$^{1}$}
\end{center}

\begin{center}
\begin{minipage}{14cm}
{
\begin{center}
{\it {\footnotesize 

\vspace{1.2mm}


$^1$ Graduate School of Informatics and Engineering,  The University of Electro-Communications, 
1-5-1 Chofugaoka, Chofu-shi, Tokyo 182-8585, Japan. \\

\vspace{1.2mm}


}}
\vspace{2mm}

yoshikawa@ai.lab.uec.ac.jp \hspace{5mm} skawano@ai.lab.uec.ac.jp \\

\end{center}


}
\end{minipage}
\end{center}

\vspace{1mm} 

\begin{abstract}
\noindent We consider the problem of constructing a reduced-rank regression model whose coefficient parameter is represented as a singular value decomposition with sparse singular vectors. The traditional estimation procedure for the coefficient parameter often fails when the true rank of the parameter is high. To overcome this issue, we develop an estimation algorithm with rank and variable selection via sparse regularization and manifold optimization, which enables us to obtain an accurate estimation of the coefficient parameter even if the true rank of the coefficient parameter is high. Using sparse regularization, we can also select an optimal value of the rank. We conduct Monte Carlo experiments and real data analysis to illustrate the effectiveness of our proposed method.
\end{abstract}

\begin{center}
\begin{minipage}{14cm}{
\vspace{3mm}
}
\end{minipage}
\end{center}

\baselineskip = 8mm


\section{Introduction}
Reduced-rank regression (RRR), a useful tool for statistics, is based on a multivariate linear regression model with a low-rank constraint for the coefficient parameter. RRR reduces the number of parameters included in the model and enables us to easily interpret the relationship between response and predictor variables. Therefore, RRR is used in various fields of research, including genomics, signal processing, and econometrics. To date, various extensions for RRR have been proposed: high-dimensional RRR with a rank selection criterion (Bunea et al., 2011), RRR with a nuclear norm penalization (Yuan et al., 2007; Negahban and Wainwright, 2011), reduced-rank ridge regression and its kernel extensions (Mukherjee and Zhu, 2011), and reduced-rank stochastic regression with sparse singular value decomposition (Chen et al., 2013).

In recent years, the number of response and predictor variables has been increasing. This causes difficulty in the estimating of parameters when the sample size is smaller than the number of the parameters included in the model. One approach for overcoming this problem is to apply a regularization method. During previous decades, sparse regularization methods, such as lasso (Tibshirani, 1996), has been the focus of attention, because they can estimate parameters and exclude irrelevant variables simultaneously. Various studies have considered a multivariate linear regression model with some sparse regularization (see, e.g., Rothman et al. (2010); Peng et al. (2010); Li et al. (2015)). Co-sparse factor regression (SFAR; Mishra et al. (2017)) was proposed in one such study. SFAR is based on both RRR and a factor analysis model by assuming that the coefficient parameter can be decomposed by singular value decomposition with both a low-rank constraint and sparsity for the singular vectors. For the estimation of parameters, Mishra et al. (2017) proposed the sequential factor extraction via co-sparse unit-rank estimation (SeCURE) algorithm. The SeCURE algorithm sequentially estimates the parameters with orthogonality and sparsity for each factor. However, the SeCURE algorithm fails to estimate the parameters when the number of latent factors is large, because the algorithm is a greedy estimation method based on the classical Gram-Schmidt orthogonalization algorithm and it is well known that the classical method does not guarantee that the optimal solution will be obtained (Bj{\"o}rck, 1967).

To overcome this problem, we propose a factor extraction algorithm with rank and variable selection via sparse regularization and manifold optimization (RVSManOpt). Manifold optimization has demonstrated excellent performance over decades of study (Bak{\i}r al., 2004; Mishra et al., 2013; Tan et al., 2019). The minimization problem of the SFAR model can be reformulated in terms of manifold optimization. Manifold optimization enables us to solve the minimization problem by taking the geometric structure of the SFAR model into consideration. By estimating the parameters on the manifold, we simultaneously obtain all latent factors. In addition, in order to select the optimal value of the rank, we introduce a regularizer which induces a hard-thresholding operator.

The remainder of the paper is organized as follows. In Section 2, we introduce RRR and derive the SFAR model from the factor regression model. In Section 3, we reformulate the minimization problem of the SFAR model based on manifold optimization. In Section 4, we provide the estimation algorithm based on manifold optimization and discuss the selection of tuning parameters. In Sections 5, Monte Carlo experiments and real data analysis support the efficacy of RVSManOpt. Concluding remarks which summarize our study are presented in Section 6. Supplementary materials and source codes of our proposed method are available at \url{https://github.com/yoshikawa-kohei/RVSManOpt}.

\section{Preliminaries}
Suppose that we obtain $n$ independent observations $\left\{ (\vec{y}_i, \vec{x}_i); i=1,\dots,n \right\}$, where $\vec{y}_i = \trans{\left[ y_{i1}, \ldots, y_{iq} \right]} \in \mathbb{R}^{q}$ is a $q$-dimensional vector of response variables and $\vec{x}_i =  \trans{\left[ x_{i1}, \ldots, x_{ip} \right]} \in \mathbb{R}^p$ is a $p$-dimensional vector of predictor variables. When we set $\mat{Y} = \trans{\left[ \vec{y}_1, \ldots, \vec{y}_n \right]} \in \mathbb{R}^{n \times q}$ and $\mat{X} = \trans{\left[ \vec{x}_1, \ldots, \vec{x}_n \right]} \in \mathbb{R}^{n \times p}$, RRR (Anderson, 1951; Izenman, 1975; Reinsel and Velu, 1998) is formulated as
   \begin{align}
    \label{eq:RRR_model}
      \mat{Y} = \mat{X}\mat{C} + \mat{E},\quad \st \rank{{\mat{C}}} \leq r,
  \end{align}
  where $\mat{C} \in \mathbb{R}^{p \times q}$ is the coefficient matrix, which has rank at most $r = \min \left(\rank{\mat{X}}, q \right)$, and $\mat{E} = \trans{\left[ \vec{e}_1, \ldots, \vec{e}_n \right]} \in \mathbb{R}^{n \times q}$ is the error matrix, which consists of independent random error vectors $\vec{e}_i$ with mean $\mathrm{E}\left[ \vec{e}_i\right] = \vec{0}$ and covariance matrix $\mathrm{Cov}\left[ \vec{e}_i \right] = \mat{\Sigma} \ (i = 1,\dots,n)$. The estimator of the coefficient matrix $\mat{C}$ can be obtained by solving the minimization problem
  \begin{equation}
    \label{eq:min_prob_RRR_model}
      \min_{\mat{C}}\ \norm{\mat{Y} - \mat{X} \mat{C}}_F^2, \st \rank{\mat{C}}  \leq r,
  \end{equation}
  where $\norm{\cdot}_F$ denotes the Frobenius norm.

  Mishra et al. (2017) proposed SFAR by extending RRR in terms of factor analysis. Before introducing SFAR, we describe the relationship between RRR and factor analysis. First, we consider the RRR model with a coefficient matrix $\mat{C}$ that is decomposed as
\begin{align}
    \mat{C} = \mat{U} \trans{{\tilde{\mat{V}}}},
\end{align}
where $\mat{U} \in \mathbb{R}^{p \times r}$ and $\tilde{\mat{V}} \in \mathbb{R}^{q \times r}$. Then we obtain the RRR model reformulated by
\begin{align}
    \label{eq:RRR_reformulated}
    \mat{Y} = \mat{X} \mat{U} \trans{{\tilde{\mat{V}}}} + \mat{E}.
\end{align}
The equation (\ref{eq:RRR_reformulated}) is related to a factor analysis model: $\mat{XU}$ can be regarded as a common factor matrix and $\tilde{\mat{V}}$ can be regarded as a loading matrix. Furthermore, if we assume $\mathrm{E}[\vec{x}_i] = \vec{0}$ and $\mathrm{cov}[\vec{x}_i] = \mat{\Gamma}_i \ (i=1,\dots,n)$, then $\mathrm{cov}[\trans{{\mat{U}}} \vec{x}] = \trans{{\mat{U}}} \mat{\Gamma} \mat{U} = \mat{I}_r$ is derived. This in turn gives the following SFAR model.
\begin{align}
    \mat{Y} = \mat{X} \mat{U} \mat{D} \trans{{\mat{V}}} + \mat{E}, \st \trans{{\mat{U}}} \mat{\Gamma} \mat{U} = \mat{I}_{r}, \trans{{\mat{V}}} \mat{V} = \mat{I}_{r}.
\end{align}
Here, the coefficient matrix is $\mat{C} = \mat{U} \mat{D} \trans{{\mat{V}}}$.

The estimator of SFAR is obtained by solving the minimization problem
\begin{align}
    \label{eq:co-sparse_factor_regression}
    \min_{\mat{U}, \mat{D},\mat{V}}\  \frac{1}{2} \norm{\mat{Y} - \mat{X} \mat{U} \mat{D} \trans{{\mat{V}}}}_F^2 + \lambda_1 \sum_{i=1}^{p} \sum_{j=1}^{r} w^{(u)}_{ij}|{u}_{ij}| +  \lambda_2 \sum_{i=1}^{q} \sum_{j=1}^{r} w^{(v)}_{ij}|{v}_{ij}|, \nonumber \\
    \st \trans{{\mat{U}}} \left(\frac{\trans{\mat{X}} \mat{X}}{n} \right) \mat{U}= \mat{I}_{r}, \trans{{\mat{V}}} \mat{V} = \mat{I}_{r},
\end{align}
where $u_{ij}, v_{ij}$ are elements of $\mat{U}$ and $\mat{V}$, respectively, $w^{(u)}_{ij}, w^{(v)}_{ij}$ are adaptive weights with positive values proposed by Zou (2006), and $\lambda_1, \lambda_2 > 0$ are regularization parameters. The second and third terms are penalty functions inducing elementwise sparsity (Tibshirani, 1996). By solving this minimization problem, we obtain the estimator of the coefficient matrix $\hat{\mat{C}} = \hat{\mat{U}} \hat{\mat{D}} \trans{\hat{{\mat{V}}}}$.

The minimization problem is solved under orthogonality and sparsity of the parameters. However, it is difficult to estimate the parameters directly. For this reason, Mishra et al. (2017) proposed the SeCURE algorithm. The SeCURE algorithm sequentially solves the minimization problem for the $k$-th latent factor given by
\begin{align}
    \label{eq:Q_problem}
    \min_{d_k, \vec{u}_k, \vec{v}_k} \frac{1}{2} \norm{\mat{Y}_k - d_k \mat{X} \vec{u}_k \trans{\vec{v}_k}}_F^2 + \sum_{i = 1}^p w_{ki}^{(u)} |u_{ki}| + \sum_{i = 1}^q w_{ki}^{(v)} |v_{ki}|, \nonumber\\
    \st d_k \geq 0, \trans{\vec{u}_k} \trans{\mat{X}} \mat{X} \vec{u}_k = n, \trans{\vec{v}_k}\vec{v}_k = 1,
\end{align}
  where $k=1,\dots,r$, $\vec{u}_k$ and $\vec{v}_k$ are the $k$-th column vector of $\mat{U}$ and $\mat{V}$, respectively, and $\mat{Y}_k$ is defined by
\begin{align}
\mat{Y}_k = \mat{Y} - \sum_{j=1}^{k-1} d_j \mat{X} \vec{u}_j \trans{{\vec{v}_j}},
\end{align}
  in which $d_j$ is the $j$-th diagonal element of $\mat{D}$ and $\mat{Y}_1 = \mat{Y}$. By sequentially solving the minimization problem (\ref{eq:Q_problem}), we obtain the solutions $\hat{d}_k, \hat{\vec{u}}_k$, and $\hat{\vec{v}}_k$ which satisfy orthogonality and sparsity. When $\hat{\vec{u}}_k = \vec{0}$ or $\hat{\vec{v}}_k = \vec{0}$, the SeCURE algorithm updates $d_k = 0$. This means that the updates are terminated. In addition, the index $k$ that terminates the updates is regarded as the optimal value of the rank of the coefficient matrix $\mat{C}$. It should be noted that the estimation method for the minimization problem (\ref{eq:Q_problem}) is the block coordinate descent algorithm proposed by Chen et al. (2012).

\section{Minimization problem of co-sparse factor regression via manifold optimization}
The SeCURE algorithm fails to estimate the parameters for the $k$-th latent factor when $k$ is large, because the algorithm is based on the classical Gram-Schmidt orthogonalization algorithm. Note that the classical Gram-Schmidt orthogonalization algorithm does not produce an optimal solution, owing to rounding errors (Bj{\"o}rck, 1967). To overcome this problem, we reconsider this minimization problem in terms of manifold optimization.

\subsection{Reformulation of the minimization problem as manifold optimization}
To consider the minimization problem (\ref{eq:co-sparse_factor_regression}) in terms of manifold optimization, we use the fundamental geometric structure given by
\begin{align}
    \label{eq:Stiefel}
    \St{r}{q} &:= \left\{ \mat{V} \in \mathbb{R}^{q \times r} \mid \trans{\mat{V}} \mat{V} = \mat{I}_r \right\},
\end{align}
where $q \geq r$. Here, $\St{r}{q}$ is called the Stiefel manifold, which is the set of orthogonal matrices of size $q \times r$. Furthermore, we also use the generalized Stiefel manifold given by
\begin{align}
    \label{eq:gen_Stiefel}
    \StG{r}{p} &:= \left\{ \mat{U} \in \mathbb{R}^{p \times r} \mid \trans{\mat{U}} \mat{G} \mat{U} = \mat{I}_r \right\},
\end{align}
where $p \geq r$ and $\mat{G} \in \mathbb{R}^{p \times p}$ is a symmetric positive definite matrix. In this paper, we use $\mat{G} = \trans{\mat{X}} \mat{X} /n$.

By utilizing the geometric structures (\ref{eq:Stiefel}) and (\ref{eq:gen_Stiefel}), the minimization problem (\ref{eq:co-sparse_factor_regression}) can be reformulated as
\begin{align}
    \label{eq:riemannian-co-sparse-factor-regression}
    \min_{\substack{ \mat{U} \in \StG{r}{p},\\ \mat{D}\in \mathbb{R}^{r \times r},\\ \mat{V}\in \St{r}{q} }}   \frac{1}{2} \norm{\mat{Y} - \mat{X} \mat{U} \mat{D} \trans{{\mat{V}}}}_F^2 + n\lambda_1 \sum_{i=1}^{p} \sum_{j=1}^{r} w^{(u)}_{ij}|{u}_{ij}| +  n\lambda_2 \sum_{i=1}^{q} \sum_{j=1}^{r} w^{(v)}_{ij}|{v}_{ij}|.
\end{align}
The minimization problem (\ref{eq:riemannian-co-sparse-factor-regression}) is an unconstrained optimization problem, and solving it allows us to estimate all the parameters for all the latent factors at once.

\subsection{Rank selection with sparse regularization}
The reformulation of the minimization problem (\ref{eq:co-sparse_factor_regression}) gives us the unconstrained optimization problem (\ref{eq:riemannian-co-sparse-factor-regression}).
However, we cannot select the optimal value of the rank of the coefficient matrix $\mat{C}$ because of not using a sequential estimating procedure, such as SeCURE. To overcome this drawback, we propose the following minimization problem:
\begin{multline}
    \label{eq:riemannian-co-sparse-factor-regression-rank-select}
    \min_{\substack{ \mat{U} \in \StG{r}{p},\\ \mat{D}\in \mathbb{R}^{r \times r},\\ \mat{V}\in \St{r}{q} }} \frac{1}{2} \norm{\mat{Y} - \mat{X} \mat{U} \mat{D} \trans{\mat{V}}}_F^2 + n\lambda_1 \sum_{i=1}^{p} \sum_{j=1}^{r} w^{(u)}_{ij}|{u}_{ij}| \\
    + n\alpha \lambda_2 \sum_{i=1}^{q} \sum_{j=1}^{r} w^{(v)}_{ij}|{v}_{ij}| + n\sqrt{q}(1-\alpha)\lambda_2 \sum_{i=1}^r w^{(d)}_{i} \mathds{1} (\vec{v}_i \neq \vec{0}),
\end{multline}
where $\mathds{1} ( \cdot )$ is an indicator function that returns $1$ if the condition is true and returns $0$ if the condition is false, $w^{(d)}_{i}$ is an adaptive weight with a positive value proposed by Zou (2006), and $\alpha$ is a tuning parameter having a value between zero and one. The group selection in the fourth term plays the role of the rank selection of the coefficient matrix $\mat{C}$. The tuning parameter $\alpha$ adjusts the trade-off between the third term and the fourth term. The two terms can be regarded as Sparse Group Lasso (Wu and Lange, 2008; Puig et al., 2009; Simon et al., 2013). The fourth term is a regularizer which induces a hard-thresholding operator. By imposing this regularization, we can estimate some column vectors of $\mat{V}$ as zero vectors. As a consequence, the model is constructed with a small number of latent factors. In that sense, the indicator function plays the role of selecting the rank of the coefficient matrix $\mat{C}$. The reason why we do not apply Group Lasso, which induces a soft-thresholding operator (Yuan and Lin, 2006), is to avoid a double shrinking effect for the parameter $\mat{V}$. If we assume that the fourth term corresponds to the Group Lasso, then such a double shrinking effect appears to occur. The double shrinking effect reduces the variance of the model, but it excessively increases the bias. To prevent the double shrinking effect for the parameter $\mat{V}$, we use a regularizer which induces a hard-thresholding operator, since it does not shrink the value of the parameter.

\section{Implementation}
\subsection{Computational algorithm}
To estimate the parameters, we employ a manifold optimization method (Edelman et al., 1998; Absil et al., 2008). Manifold optimization can be performed for differentiable functions. However, the minimization problem (\ref{eq:riemannian-co-sparse-factor-regression-rank-select}) includes nondifferentiable penalty terms. For this reason, we handle the nondifferentiability by applying the manifold alternating direction method of multipliers (M-ADMM) proposed by Kovnatsky et al. (2016) to the minimization problem (\ref{eq:riemannian-co-sparse-factor-regression-rank-select}).

Letting $\mat{U}^* \in \mathbb{R}^{p \times r}$ and $\mat{V}^*$ and $\mat{V}^{**} \in \mathbb{R}^{q \times r}$ denote variables for splitting nondifferentiable penalty terms from the minimization problem (\ref{eq:riemannian-co-sparse-factor-regression-rank-select}), we consider a minimization problem with equality constraints as follows:
\begin{multline}
    \label{eq:riemannian-co-sparse-factor-regression-equaliity}
    \min_{\substack{ \mat{U} \in \StG{r}{p},\\ \mat{D}\in \mathbb{R}^{r \times r},\\ \mat{V}\in \St{r}{q} }} \frac{1}{2} \norm{\mat{Y} - \mat{X} \mat{U} \mat{D} \trans{\mat{V}}}_F^2 + n\lambda_1 \sum_{i=1}^{p} \sum_{j=1}^{r} w^{(u)}_{ij}|{u}_{ij}^*| \\
    + n\alpha \lambda_2 \sum_{i=1}^{q} \sum_{j=1}^{r} w^{(v)}_{ij}|{v}_{ij}^*| + n\sqrt{q}(1-\alpha)\lambda_2 \sum_{i=1}^r w^{(d)}_{i} \mathds{1} (\vec{v}_i^{**} \neq \vec{0}) ,\\
     \st \mat{U} = \mat{U}^*,\ \mat{V} = \mat{V}^* = \mat{V}^{**},
\end{multline}
where $u_{ij}^*, v_{ij}^*$ are the $(i,j)$-th elements of $\mat{U}^*$ and $\mat{V}^*$, respectively, and $\vec{v}_i^{**}$ is an $i$-th column vector of $\mat{V}^{**}$. When we let $\mat{\Omega} \in \mathbb{R}^{p \times r}$ and $\mat{\Phi}$ and $\mat{\Psi} \in \mathbb{R}^{q \times r}$ denote the dual variables, we obtain a scaled augmented Lagrangian (Boyd et al., 2011) as follows:
\begin{multline}
    \mathcal{L}(\mat{U}, \mat{D}, \mat{V}, \mat{U}^*, \mat{V}^*, \mat{V}^{**}, \mat{\Omega}, \mat{\Phi}, \mat{\Psi}) = \frac{1}{2} \norm{\mat{Y} - \mat{X} \mat{U} \mat{D} \trans{\mat{V}}}_F^2\\
    + n\lambda_1 \sum_{i=1}^{p} \sum_{j=1}^{r} w^{(u)}_{ij}|u_{ij}^*| + n\alpha \lambda_2 \sum_{i=1}^{q} \sum_{j=1}^{r} w^{(v)}_{ij}|v_{ij}^*| + n\sqrt{q}(1-\alpha)\lambda_2 \sum_{i=1}^r w^{(d)}_{i} \mathds{1} (\vec{v}_i^{**} \neq \vec{0}) \\
    + \frac{\rho_1}{2} \norm{\mat{U} - \mat{U}^* + \mat{\Omega}}_F^2  + \frac{\rho_2}{2} \norm{\mat{V} - \mat{V}^* + \mat{\Phi}}_F^2 + \frac{\rho_3}{2} \norm{\mat{V} - \mat{V}^{**} + \mat{\Psi}}_F^2,
\end{multline}
where $\rho_1, \rho_2, \rho_3 > 0 $ are penalty parameters. For this study, we fixed $\rho_1 = \rho_2 = \rho_3 = 1$. M-ADMM alternately updates each parameter to minimize the augmented Lagrangian. The estimators of elements in $\mat{U}^*$ and $\mat{V}^*$ indicate whether each element of the parameter is zero. The estimators of column vectors in $\mat{V}^{**}$ indicate whether each vector of the parameter is a zero vector. In the M-ADMM procedure, we initialize the parameters by using $\tilde{\mat{U}} \in \mathbb{R}^{p \times r}, \tilde{\mat{D}} = \mathrm{diag} (\tilde{d}_1, \dots, \tilde{d}_r), \tilde{\mat{V}} \in \mathbb{R}^{q \times r}$.
Here, $\tilde{\mat{U}}$ is calculated by $(\trans{\mat{X}} \mat{X})^{-} \trans{\mat{X}} \mat{Y} \tilde{\mat{V}} \tilde{\mat{D}}^{-1}$, where the $k$-th diagonal element of $\tilde{\mat{D}}^2$ is the $k$-th eigenvalue of $(1/n) \trans{\mat{Y}} \mat{X} (\trans{\mat{X}} \mat{X})^{-} \trans{\mat{X}}\mat{Y}$, and the $k$-th column vector of $\tilde{\mat{V}}$ is the $k$-th eigenvalue of $(1/n) \trans{\mat{Y}} \mat{X} (\trans{\mat{X}} \mat{X})^{-} \trans{\mat{X}}\mat{Y}$.

We set the adaptive weights $w_{ij}^{(u)}, w_{ij}^{(v)}, w_{i}^{(d)}$ as
\begin{align}
    w^{(u)}_{ij} &= \frac{1}{\left|\tilde{u}_{ij} \right|^{{\gamma^u}}}, \quad i=1,\dots,p, j=1,\dots,r,\\
    w^{(v)}_{ij} &= \frac{1}{\left|\tilde{v}_{ij} \right|^{{\gamma^v}}}, \quad i=1,\dots,q, j=1,\dots,r,\\
    w^{(d)}_{i} &= \frac{1}{|\tilde{d}_{i} |^{{\gamma^d}}}, \quad i=1,\dots,r,
  \end{align}
where $\gamma^u$, $\gamma^v$, $\gamma^d > 0$ are tuning parameters.

The parameters $\mat{U}$ and $\mat{V}$ are estimated by a gradient descent algorithm based on manifold optimization. For example, the procedure for estimating $\mat{U}$ can be represented by the following.
\begin{enumerate}
    \item At a given iteration $s$, calculate the Euclidean gradient $\nabla \mathcal{L}_{\mat{U}^{(s)}}$.
    \item Project $\nabla \mathcal{L}_{\mat{U}^{(s)}}$ onto the tangent space $\mathcal{T}_{\mat{U}^{(s)}} \StG{p}{r}$ using orthogonal projection $\mathcal{P}_{\mat{U}^{(s)}}(\cdot)$ to obtain the gradient $\mathrm{grad} \mathcal{L}_{\mat{U}^{(s)}}$ on the manifold.
    \item Update the parameter $\mat{U}^{(s)}$ by retraction $\mathcal{R}_{\mat{U}^{(s)}}(- t\ \mathrm{grad} \mathcal{L}_{\mat{U}^{(s)}})$ to obtain the parameter $\mat{U}^{(s+1)}$, where $t \in \mathbb{R}$ is an Armijo step size described in Absil et al. (2008).
\end{enumerate}
The necessary notation is shown in Table \ref{tb:notations_manifold}. In the same way, we estimate the parameter $\mat{V}$ on the manifold. The detailed calculation of the updates is described in the Appendix. This algorithm is called the \textit{factor extraction algorithm with rank and variable selection via sparse regularization and manifold optimization} (RVSManOpt). RVSManOpt is summarized as Algorithm \ref{alg:r-sgfar_admm}.

\begin{table}[htbp]
    \centering
    \caption{Notation for the manifold optimization algorithm}
    \begin{tabular}{ll}
    \hline
    \multicolumn{2}{c}{Generalized Stiefel manifold for parameter $\mat{U}$}\\
    Metric & $\langle \mat{U}_1, \mat{U}_2 \rangle = \Tr ( \trans{\mat{U}_1} \mat{G} \mat{U}_2), \mat{G} = \trans{\mat{X}}\mat{X}/n$\\
    Tangent space & $\mathcal{T}_\mat{U} \StG{p}{r} = \{ \mat{Z} \in \mathbb{R}^{p \times r} | \trans{\mat{U}} \mat{G} \mat{Z} + \trans{\mat{Z}} \mat{G} \mat{U} = \mat{0} \}$\\
    Projection onto tangent space & $\mathcal{P}_\mat{U}(\mat{Z}) = \mat{Z} - \mat{U} \mathrm{sym}(\trans{\mat{U}}\mat{G} \mat{Z}), \mathrm{sym}(\mat{M}) = \frac{1}{2} (\mat{M} + \trans{\mat{M}} )$\\
    Gradient & $\mathrm{grad} \mathcal{L}_\mat{U} = \mathcal{P}_\mat{U}(\nabla \mathcal{L}_\mat{U})$\\
    Retraction mapping & $\mathcal{R}_\mat{U}(\mat{Z})= \sqrt{\mat{G}}^{-1} \mathrm{qf} \left(\sqrt{\mat{G}}(\mat{U} + \mat{Z}) \right)$,\\
    & $\mathrm{qf}(\mat{A})$ denotes the $\mat{Q}$ factor of the QR decomposition of $\mat{A} = \mat{Q}\mat{R}$\\
    \\
    \multicolumn{2}{c}{Stiefel manifold for parameter $\mat{V}$}\\
    Metric & $\langle \mat{V}_1, \mat{V}_2 \rangle = \Tr ( \trans{\mat{V}_1} \mat{V}_2)$\\
    Tangent space & $\mathcal{T}_\mat{V} \St{q}{r} = \{ \mat{Z} \in \mathbb{R}^{q \times r} | \trans{\mat{V}} \mat{Z} + \trans{\mat{Z}} \mat{V} = \mat{0} \}$\\
    Projection onto tangent space & $\mathcal{P}_\mat{V}(\mat{Z}) = \mat{Z} - \mat{V} \mathrm{sym}(\trans{\mat{V}} \mat{Z})$\\
    Gradient & $\mathrm{grad} \mathcal{L}_\mat{V} = \mathcal{P}_\mat{V}(\nabla \mathcal{L}_\mat{V})$\\
    Retraction mapping & $\mathcal{R}_\mat{V}(\mat{Z}) = \mathrm{qf}(\mat{V} + \mat{Z})$\\
    \hline
    \end{tabular}
    \label{tb:notations_manifold}
\end{table}
\begin{spacing}{0.9}
\begin{algorithm}[H]
    \caption{Factor Extraction Algorithm with Rank and Variable Selection via Sparse Regularization and Manifold Optimization (RVSManOpt)}
    \label{alg:r-sgfar_admm}
    \begin{algorithmic}[1]
        \Require Initial values $\mat{U}^{(0)} = \tilde{\mat{U}}, \mat{D}^{(0)} = \tilde{\mat{D}} , \mat{V}^{(0)} = \tilde{\mat{V}},
        {\mat{U}^* }^{(0)}= \mat{U}^{(0)}, {\mat{V}^{*}}^{(0)} ={\mat{V}^{**}}^{(0)} =  \mat{V}^{(0)}, \mat{\Omega}^{(0)} = \mat{0}, \mat{\Phi}^{(0)}, \mat{\Psi}^{(0)} = \mat{0}$
        \For{$s=0,1,\ldots$}
            \State \textbf{$\mat{U}$ Step:} Update $\mat{U}^{(s+1)} \gets \mathcal{R}_{\mat{U}^{(s)}}(-t_u^{(s)} \mathrm{grad} \mathcal{L}_{\mat{U}^{(s)}})$, $t_u^{(s)}$ is the Armijo step size.

            \State \textbf{$\mat{V}$ Step:} Update $\mat{V}^{(s+1)} \gets \mathcal{R}_{\mat{V}^{(s)}}(-t_v^{(s)} \mathrm{grad} \mathcal{L}_{\mat{V}^{(s)}})$, $t_v^{(s)}$ is the Armijo step size.

            \State \textbf{$\mat{D}$ Step:} Update $\mat{D}^{(s+1)} \gets \mathrm{diag}\left( \frac{1}{n}\trans{{\mat{V}^{(s+1)}}} \trans{\mat{Y}} \mat{X} \mat{U}^{(s+1)} \right)$.

            \State \textbf{$\mat{U}^*$ Step:}

            \For{$i=1,\ldots,p$}
                \For{$j=1,\ldots,r$}
                    \State Update ${u_{ij}^*}^{(s+1)} \gets \mathrm{S} \left(u_{ij}^{(s+1)} + \omega_{ij}^{(s)}, \frac{n\lambda_1 w_{ij}^{(u)}}{\gamma_1} \right)$.
                \EndFor
            \EndFor

            \State \textbf{$\mat{V}^*$ Step:}

            \For{$i=1,\ldots,q$}
                \For{$j=1,\ldots,r$}
                    \State Update ${v_{ij}^{*}}^{(s+1)} \gets \mathrm{S} \left(v_{ij}^{(s+1)} + \phi_{ij}^{(s)}, \frac{n\alpha \lambda_2 w_{ij}^{(v)}}{\gamma_2} \right)$.
                \EndFor
            \EndFor
            \State \textbf{$\mat{V}^{**}$ Step:}

            \For{$i=0,1,\ldots,r$}
                \State Update ${\vec{v}^{**}_{i}}^{(s+1)} \gets \mathrm{H} \left({\vec{v}}_{i}^{(s+1)} +  \vec{\psi}_{i}^{(s)}, \sqrt{\frac{2n\sqrt{q}(1-\alpha)\lambda_2 w_i^{(d)}}{\gamma_3}} \right)$.
            \EndFor

            \State \textbf{$\mat{\Omega}$ Step:} Update $\mat{\Omega}^{(s+1)} \gets \mat{\Omega}^{(s)} +  \mat{U}^{(s+1)} - {\mat{U}^*}^{(s+1)}$.

            \State \textbf{$\mat{\Phi}$ Step:} Update $\mat{\Phi}^{(s+1)} \gets \mat{\Phi}^{(s)} +  \mat{V}^{(s+1)} - \mat{{V}^*}^{(s+1)} $.

            \State \textbf{$\mat{\Psi}$ Step:} Update $\mat{\Psi}^{(s+1)} \gets \mat{\Psi}^{(s)} +  \mat{V}^{(s+1)} - \mat{{V}^{**}}^{(s+1)} $.

            \If{$\mathrm{convergence}$}
                \State \textbf{break}.
            \EndIf
        \EndFor

        \State $\hat{\mat{U}} \gets \mat{U}^*; \hat{\mat{U}} \gets \hat{\mat{U}}(:, \vec{v}_i^{**} \neq \vec{0})$, $\hat{\mat{D}} \gets \mat{D}(:, \vec{v}_i^{**} \neq \vec{0})$, $\hat{\mat{V}} \gets \mat{V}^*; \hat{\mat{V}} \gets \hat{\mat{V}}(:, \vec{v}_i^{**} \neq \vec{0})$
        \State \textbf{return} $\hat{\mat{U}}, \hat{\mat{D}}, \hat{\mat{V}}$
    \end{algorithmic}
\end{algorithm}
\end{spacing}

\subsection{Selection of tuning parameters}
We have six tuning parameters: $\lambda_1, \lambda_2, \alpha, \gamma^u, \gamma^v$, and $\gamma^d$. To avoid a high computational cost, $\alpha, \gamma^u, \gamma^v$, and $\gamma^d$ are fixed in advance. We set the values of these tuning parameters according to the situation. The tuning parameter $\alpha$ is set to a large value when a sparse regularization is more important than a regularization for selecting the rank of the coefficient matrix $\mat{C}$. Larger values of tuning parameters $\gamma^u, \gamma^v$, and $\gamma^d$ correspond to a higher data dependence. To select the remaining two tuning parameters, $\lambda_1$ and $\lambda_2$, we use the Bayesian information criterion (BIC) given by
\begin{align}
    \label{eq:BIC}
    \mathrm{BIC} = \log \left\{ \mathrm{SSE}_{\lambda_1, \lambda_2} / nq \right\} + \left\{ \log(qn)/(nq) \right\} df_{\lambda_1, \lambda_2},
\end{align}
where $\mathrm{SSE}_{\lambda_1, \lambda_2}$ is the sum of squared errors of prediction defined by
\begin{equation}
    \mathrm{SSE}_{\lambda_1, \lambda_2} = \norm{\mat{Y} - \mat{X} \hat{\mat{U}} \hat{\mat{D}} \trans{\hat{\mat{V}}}}_F^2,
  \end{equation}
and $df_{\lambda_1, \lambda_2}$ is the degree of freedom which evaluates the sparsity of the estimates $\hat{\mat{U}}$ and $\hat{\mat{V}}$ defined by
\begin{equation}
    df_{\lambda_1, \lambda_2} = \sum_{i=1}^{p} \sum_{j=1}^r \mathds{1}(\hat{u}_{ij} \neq 0) + \sum_{i=1}^{q} \sum_{j=1}^r \mathds{1}(\hat{v}_{ij} \neq 0)  - 1.
  \end{equation}
We select the tuning parameters $\lambda_1$ and $\lambda_2$ which minimize the BIC. The candidates values of $\lambda_1, \lambda_2$ are taken from equally spaced values in the interval $[\lambda_{\max}, \lambda_{\min}]$. We set $\lambda_{\max}=1$ and $\lambda_{\min}=10^{-15}$ in our numerical studies.

\section{Numerical study}
\subsection{Monte Carlo simulations}
We conducted Monte Carlo simulations to illustrate the efficacy of RVSManOpt. In our simulation study, we generated 50 datasets from the model:
\begin{align}
    \mat{Y} = \mat{XC} + \mat{E},
\end{align}
where $\mat{Y} \in \mathbb{R}^{n \times q}$ is a response matrix, $\mat{X} \in \mathbb{R}^{n \times p}$ is a predictor matrix, $\mat{C} \in \mathbb{R}^{p \times q}$ is a coefficient matrix, and $\mat{E} = \trans{[\vec{e}_1, \dots \vec{e}_n]} \in \mathbb{R}^{n \times q}$ is an error matrix. Each row of $\mat{X}$ followed a multivariate normal distribution $\mathcal{N}(\vec{0}, \mat{\Gamma})$, where $\mat{\Gamma} = [\gamma_{ij}]$ is a $p \times p$ covariance matrix with $\gamma_{ij} = 0.5^{|i-j|}$ for $i,j = 1,\dots,p$. We generated each row of $\mat{E}$ by $\vec{e}_i \overset{\mathrm{i.i.d.}}{\sim} \mathcal{N}(\vec{0}, \sigma^2 \mat{\Delta})$, where $\mat{\Delta} = [\delta_{ij}]$ is a $q \times q$ matrix with $\delta_{ij} = \rho^{|i-j|}$ and $\sigma$ is determined according to the signal-to-noise ratio defined by $\mathrm{SNR} = \norm{d_r \mat{X} \vec{u}_r \trans{\vec{v}}}_2 / \norm{\mat{E}}_2 = 0.5$. We considered the ranks of the coefficient matrix as follows: $r \in \{3, 5, 7, 10, 12 \}$. We generated the coefficient matrix $\mat{C} = \mat{UD}\trans{\mat{V}}$, where $\mat{U} = [\vec{u}_1, \dots, \vec{u}_r]$, $\mat{D} = \mathrm{diag}(d_1, \dots, d_r)$, $\mat{V} = [\vec{v}_1, \dots, \vec{v}_r]$. Specifically, we set
\begin{align*}
    d_k &= 5 + 0.1(k-1), \quad k = 1, \dots ,r,\\
    \vec{u}_k &= \bar{\vec{u}}_k/\norm{\bar{\vec{u}}_k}_2,\\
    \bar{\vec{u}}_1 &= \trans{[\check{\vec{u}}, \mathrm{rep}(0, p-8)]}, \bar{\vec{u}}_k = \trans{[\mathrm{rep}(0,5(k-1)),\check{\vec{u}}, \mathrm{rep}(0, p-(5k+3))]},\\
     \check{\vec{u}} &= [1, -1, 1, -1, 0.5, -0.5, 0.5, -0.5],\\
     \vec{v}_k &= \bar{\vec{v}}_k/\norm{\bar{\vec{v}}_k}_2,\\
     \bar{\vec{v}}_1 &= \trans{[\check{\vec{v}}, \mathrm{rep}(0, q-4)]}, \bar{\vec{v}}_k = \trans{[\mathrm{rep}(0,4(k-1)), \check{\vec{v}}, \mathrm{rep}(0, q-4k)]},\\
      \check{\vec{v}} &= [1, -1, 0.5, -0.5],
\end{align*}
where $\mathrm{rep}(a, b)$ represents the vector of length $b$ with all elements having the value $a$. We considered four cases. In Cases 1 and 2, we set $n=400, p=80$, and $q=50$ in common, and we set the correlation as $\rho=0.3$ (Case 1) or $\rho=0.5$ (Case 2). In Cases 3 and 4, we set $n=400, p=120$, and $q=60$ in common, and we set the correlation as $\rho=0.3$ (Case 3) or $\rho=0.5$ (Case 4).

To demonstrate the efficacy of RVSManOpt, we compared RVSManOpt with the SeCURE with an adaptive lasso (SeCURE(AL)), and the SeCURE with an adaptive elastic net (SeCURE(AE)). For 50 datasets, we measured the estimation accuracy $\mathrm{Er}(\mat{XC})
$ and the selected rank absolute error $\mathrm{Er}(r)$. These are defined as
\begin{align}
    \mathrm{Er}(\mat{XC}) &= \frac{1}{50} \sum_{k=1}^{50}\frac{\norm{\mat{\Gamma}^{\frac{1}{2}}(\hat{\mat{C}}^{(k)} - \mat{C}^{(k)})}_F^2}{nq},\\
    \mathrm{Er}(r) &= \frac{1}{50} \sum_{k=1}^{50}|\hat{r}^{(k)} - r^{(k)}|,
\end{align}
where $\mat{C}^{(k)}$ is the true coefficient matrix, $r^{(k)}$ is the true rank of the coefficient matrix $\mat{C}^{(k)}$, $\hat{\mat{C}}^{(k)}$ is an estimated coefficient matrix, and $\hat{r}^{(k)}$ is the selected rank of coefficient matrix $\hat{\mat{C}}^{(k)}$ for the $k$-th dataset. In order to evaluate the sparsity, we computed the F-measure defined by
\begin{align*}
    \textrm{F-measure} &= \frac{1}{50} \sum_{k=1}^{50} 2 \cdot \frac{\mathrm{Recall}^{(k)} \cdot \mathrm{Precision}^{(k)}}{\mathrm{Recall}^{(k)} + \mathrm{Precision}^{(k)}},
\end{align*}
where $\mathrm{Recall}^{(k)}$ and $\mathrm{Precision}^{(k)}$ are defined by
\begin{align*}
    \textrm{Recall}^{(k)} &=  \frac{ \sum_{ij} \left| \left\{ u_{ij} \neq 0 \wedge \hat{u}_{ij}^{(k)} \neq 0 \right\} \right| }{  \sum_{ij} \left| \left\{u_{ij} \neq 0\right\}  \right| } + \frac{ \sum_{ij} \left| \left\{ v_{ij} \neq 0 \wedge \hat{v}_{ij}^{(k)} \neq 0 \right\} \right| }{  \sum_{ij} \left| \left\{v_{ij} \neq 0\right\}  \right| },\\
    \textrm{Precision}^{(k)} &=  \frac{ \sum_{ij} \left| \left\{ u_{ij} \neq 0 \wedge \hat{u}_{ij}^{(k)} \neq 0 \right\} \right| }{  \sum_{ij}  \left| \left\{\hat{u}_{ij}^{(k)} \neq 0\right\}  \right|} + \frac{ \sum_{ij} \left| \left\{ v_{ij} \neq 0 \wedge \hat{v}_{ij}^{(k)} \neq 0 \right\} \right| }{  \sum_{ij}  \left| \left\{\hat{v}_{ij}^{(k)} \neq 0\right\}  \right|},
\end{align*}
for which $\hat{u}_{ij}^{(k)}$ and $\hat{v}_{ij}^{(k)}$ are respectively elements of the estimated $\mat{U}$ and $\mat{V}$ for the $k$-th dataset and $|\{\cdot \}|$ is the count of the elements of set $\{\cdot \}$. All implementations were done in \texttt{R} (ver. 3.6) (R Core Team, 2018).

Tables \ref{tb:resuts_table1}, \ref{tb:resuts_table2}, \ref{tb:resuts_table3}, and \ref{tb:resuts_table4} show summaries of the results for, respectively, Cases 1 to 4 of the Monte Carlo simulations. As shown, when the rank of the coefficient matrix $\mat{C}$ is high, RVSManOpt outperforms other algorithms in terms of both $\mathrm{Er}(\mat{XC})$ and $\mathrm{Er}(r)$. In contrast, when the rank of the coefficient matrix $\mat{C}$ is low, the performances of all algorithm are approximately the same. Moreover, the F-measure gives almost the same value for RVSManOpt, SeCURE(AL) and SeCURE(AE). Therefore, our proposed RVSManOpt achieves performance superior to those of other methods in terms of both the estimation accuracy and rank selection.

Fig. \ref{fig:ErXC} shows box-plots of $\mathrm{Er}(\mat{XC})$ for Case 1. The box-plots for the other cases are essentially same and are available as the supplementary materials. When the rank of the coefficient matrix $\mat{C}$ is high, we observe many outliers in the box-plots of SeCURE(AL) and SeCURE(AE). These outliers indicate that SeCURE(AL) and SeCURE(AE) fail to estimate parameters many times. On the other hand, the number of the outliers produced by RVSManOpt is small, and hence RVSManOpt performs the other methods in terms of stable estimation.

\begin{table}[H]
    \begin{center}
    \caption{Results for Monte Carlo simulations in Case 1. For simplicity, $\mathrm{Er}(\mat{XC})$ is multiplied by $10^4$.}    \begin{tabular*}{\textwidth}{@{\extracolsep{\fill}} c|lcccc}
        \hline
    TrueRank & Method & $\mathrm{Er}(\mat{XC})$ & $\mathrm{Er}(\mat{XC})$(sd) & F-measure & $\mathrm{Er}(r)$\\
    \hline
    \hline
    & \multicolumn{5}{c}{Case 1 : $n = 400, p=80, q = 50, \rho= 0.3$}\\
    \multirow{3}{*}{3}  & RVSManOpt & 0.42  & 0.20  & 0.59  & 0.00 \\
     & SeCURE(AL) & 0.45  & 1.35  & 0.56  & 0.04 \\
     & SeCURE(AE) & 0.45  & 1.35  & 0.56  & 0.04 \\
     \\
     \multirow{3}{*}{5}  & RVSManOpt & 1.00  & 0.42  & 0.41  & 0.00 \\
      & SeCURE(AL) & 1.18  & 1.84  & 0.42  & 0.10 \\
      & SeCURE(AE) & 0.99  & 1.27  & 0.42  & 0.06 \\
     \\
     \multirow{3}{*}{7}  & RVSManOpt & 1.76  & 0.70  & 0.33  & 0.00 \\
     & SeCURE(AL) & 3.53  & 4.97  & 0.34  & 0.42 \\
     & SeCURE(AE) & 4.18  & 5.81  & 0.33  & 0.54 \\
     \\
     \multirow{3}{*}{10}  & RVSManOpt & 4.06  & 2.30  & 0.27  & 0.00 \\
     & SeCURE(AL) & 7.83  & 8.04  & 0.28  & 0.82 \\
     & SeCURE(AE) & 8.37  & 8.33  & 0.28  & 0.92 \\
     \\
     \multirow{3}{*}{12}  & RVSManOpt & 7.25  & 4.50  & 0.24  & 0.00 \\
     & SeCURE(AL) & 13.55  & 13.65  & 0.24  & 1.60 \\
     & SeCURE(AE) & 14.16  & 14.62  & 0.24  & 1.70 \\
    \hline
\end{tabular*}
\label{tb:resuts_table1}
\end{center}
\end{table}

\begin{figure}[H]
    \centering
    \includegraphics[width=\linewidth]{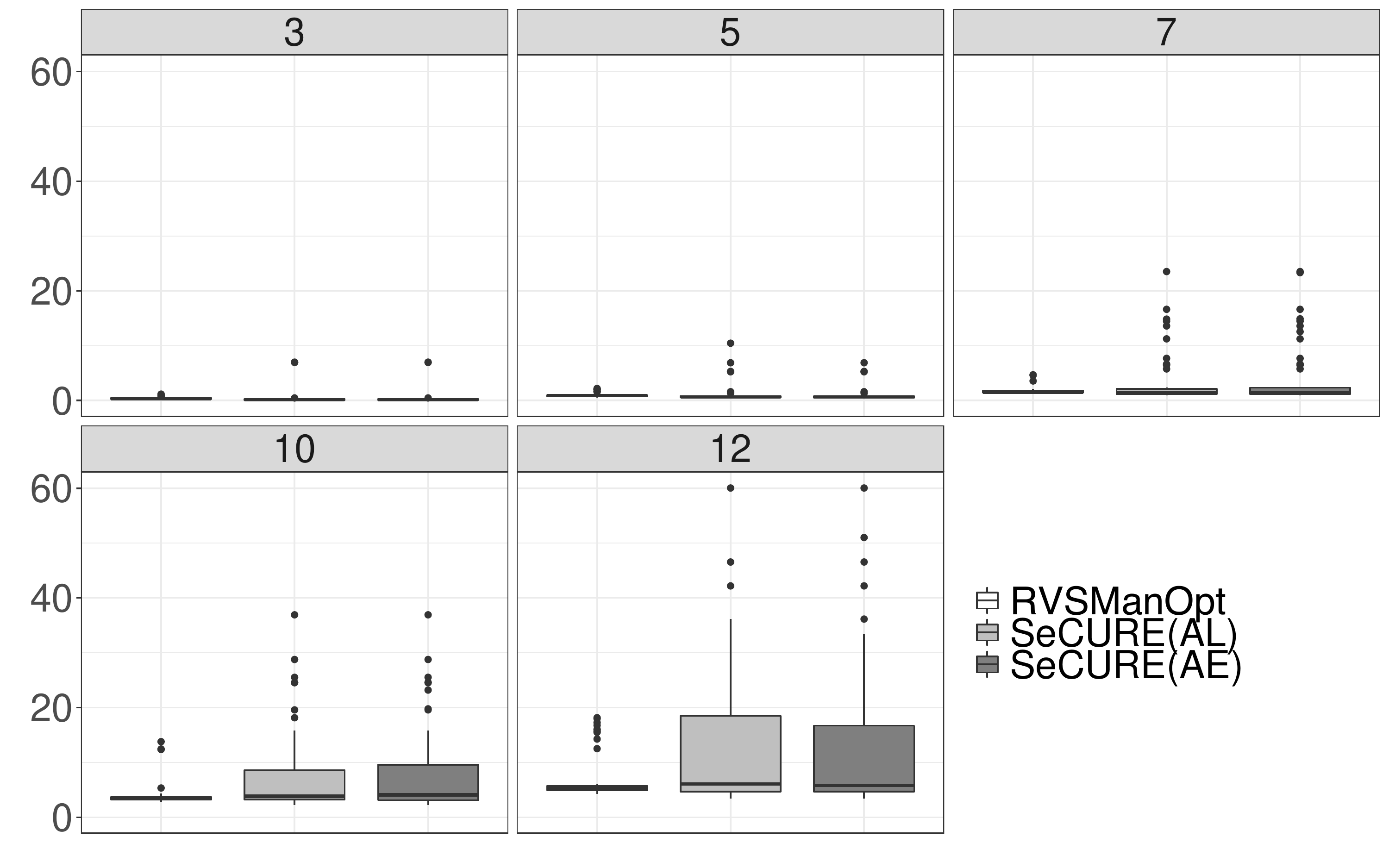}
    \caption{Box-plots of scaled $\mathrm{Er}(\mat{XC})$ for each rank $r$ in Case 1. }
    \label{fig:ErXC}
\end{figure}

\begin{table}[H]
    \begin{center}
    \caption{Results for Monte Carlo simulations in Case 2. For simplicity, $\mathrm{Er}(\mat{XC})$ is multiplied by $10^4$.}    \begin{tabular*}{\textwidth}{@{\extracolsep{\fill}} c|lcccc}
        \hline
    TrueRank & Method & $\mathrm{Er}(\mat{XC})$ & $\mathrm{Er}(\mat{XC})$(sd) & F-measure & $\mathrm{Er}(r)$\\
    \hline
    \hline
    & \multicolumn{5}{c}{Case 2 : $n = 400, p=80, q = 50, \rho= 0.5$}\\
    \multirow{3}{*}{3} & RVSManOpt & 0.38  & 0.19  & 0.57  & 0.00 \\
    &SeCURE(AL) & 0.30  & 0.98  & 0.56  & 0.02 \\
    &SeCURE(AE) & 0.30  & 0.98  & 0.56  & 0.02 \\
     \\
     \multirow{3}{*}{5} & RVSManOpt & 1.10  & 0.66  & 0.42  & 0.00 \\
     &SeCURE(AL) & 1.79  & 3.20  & 0.41  & 0.22 \\
     &SeCURE(AE) & 1.69  & 3.15  & 0.41  & 0.20 \\
     \\
     \multirow{3}{*}{7}  & RVSManOpt & 1.85  & 0.81  & 0.34  & 0.00 \\
     &SeCURE(AL) & 4.45  & 6.40  & 0.33  & 0.58 \\
     &SeCURE(AE) & 4.56  & 6.39  & 0.33  & 0.60 \\
     \\
     \multirow{3}{*}{10} & RVSManOpt & 4.39  & 2.63  & 0.27  & 0.00 \\
     &SeCURE(AL) & 8.53  & 8.62  & 0.28  & 1.00 \\
     &SeCURE(AE) & 9.33  & 9.12  & 0.28  & 1.14 \\
     \\
     \multirow{3}{*}{12} & RVSManOpt & 7.47  & 4.76  & 0.24  & 0.00 \\
     &SeCURE(AL) & 12.13  & 11.59  & 0.24  & 1.34 \\
     &SeCURE(AE) & 11.92  & 11.48  & 0.24  & 1.30 \\
    \hline
    \end{tabular*}
    \label{tb:resuts_table2}
    \end{center}
\end{table}

\begin{table}[H]
    \begin{center}
    \caption{Results for Monte Carlo simulations in Case 3. For simplicity, $\mathrm{Er}(\mat{XC})$ is multiplied by $10^4$.}    \begin{tabular*}{\textwidth}{@{\extracolsep{\fill}} c|lcccc}
        \hline
    TrueRank & Method & $\mathrm{Er}(\mat{XC})$ & $\mathrm{Er}(\mat{XC})$(sd) & F-measure & $\mathrm{Er}(r)$\\
    \hline
    \hline
    & \multicolumn{5}{c}{Case 3 : $n = 400, p=120, q = 60, \rho= 0.3$}\\
    \multirow{3}{*}{3}  & RVSManOpt & 0.40  & 0.18  & 0.56  & 0.00 \\
    &SeCURE(AL) & 0.33  & 0.94  & 0.56  & 0.04 \\
    &SeCURE(AE) & 0.33  & 0.94  & 0.56  & 0.04 \\
     \\
     \multirow{3}{*}{5}  & RVSManOpt & 0.73  & 0.21  & 0.42  & 0.00 \\
     &SeCURE(AL) & 0.84  & 1.58  & 0.41  & 0.10 \\
     &SeCURE(AE) & 1.18  & 1.88  & 0.41  & 0.18 \\
     \\
     \multirow{3}{*}{7}  & RVSManOpt & 1.77  & 1.09  & 0.34  & 0.00 \\
     &SeCURE(AL) & 3.92  & 5.63  & 0.34  & 0.64 \\
     &SeCURE(AE) & 3.81  & 5.62  & 0.34  & 0.62 \\
     \\
     \multirow{3}{*}{10}  & RVSManOpt & 3.41  & 1.50  & 0.27  & 0.00 \\
     &SeCURE(AL) & 8.87  & 9.18  & 0.27  & 1.44 \\
     &SeCURE(AE) & 9.01  & 9.27  & 0.27  & 1.46 \\
     \\
     \multirow{3}{*}{12}  & RVSManOpt & 4.59  & 2.22  & 0.23  & 0.00 \\
     &SeCURE(AL) & 12.02  & 12.44  & 0.24  & 1.84 \\
     &SeCURE(AE) & 14.93  & 14.47  & 0.24  & 2.48 \\
    \hline
\end{tabular*}
\label{tb:resuts_table3}
\end{center}
\end{table}

\begin{table}[H]
    \begin{center}
    \caption{Results for Monte Carlo simulations in Case 4. For simplicity, $\mathrm{Er}(\mat{XC})$ is multiplied by $10^4$.}    \begin{tabular*}{\textwidth}{@{\extracolsep{\fill}} c|lcccc}
        \hline
    TrueRank & Method & $\mathrm{Er}(\mat{XC})$ & $\mathrm{Er}(\mat{XC})$(sd) & F-measure & $\mathrm{Er}(r)$\\
    \hline
    \hline
    & \multicolumn{5}{c}{Case 4 : $n = 400, p=120, q = 60, \rho= 0.5$}\\
    \multirow{3}{*}{3}  & RVSManOpt & 0.42  & 0.34  & 0.57  & 0.06 \\
    &SeCURE(AL) & 0.31  & 0.94  & 0.56  & 0.04 \\
    &SeCURE(AE) & 0.31  & 0.94  & 0.56  & 0.04 \\
     \\
     \multirow{3}{*}{5}  & RVSManOpt & 1.24  & 0.75  & 0.43  & 0.08 \\
     &SeCURE(AL) & 0.96  & 1.93  & 0.41  & 0.14 \\
     &SeCURE(AE) & 1.31  & 2.19  & 0.40  & 0.22 \\
     \\
     \multirow{3}{*}{7}  & RVSManOpt & 2.28  & 1.11  & 0.35  & 0.04 \\
     &SeCURE(AL) & 4.20  & 5.27  & 0.33  & 0.72 \\
     &SeCURE(AE) & 5.14  & 6.07  & 0.32  & 0.94 \\
     \\
     \multirow{3}{*}{10}  & RVSManOpt & 4.14  & 1.99  & 0.28  & 0.10 \\
     &SeCURE(AL) & 5.92  & 5.64  & 0.27  & 0.86 \\
     &SeCURE(AE) & 7.10  & 7.08  & 0.27  & 1.10 \\
     \\
     \multirow{3}{*}{12}  & RVSManOpt & 6.46  & 3.61  & 0.24  & 0.14 \\
     &SeCURE(AL) & 13.70  & 12.58  & 0.24  & 2.22 \\
     &SeCURE(AE) & 13.49  & 12.66  & 0.24  & 2.18 \\
    \hline
\end{tabular*}
\label{tb:resuts_table4}
\end{center}
\end{table}

\subsection{Application to yeast cell cycle dataset}
We applied RVSManOpt to yeast cell cycle data (Spellman et al., 1998). The dataset was available in the \texttt{secure} package (Mishra et al., 2017) in the software \texttt{R}. The analysis of the yeast cell cycle enables us to identify transcription factors (TFs) which regulate ribonucleic acid (RNA) levels within the eukaryotic cell cycle. The dataset contains two components: the chromatin immunoprecipitation (ChIP) data and eukaryotic cell cycle data. The binding information of a subset of 1790 genes and 113 TFs was included in the ChIP data (Lee et al., 2002). The cell cycle data were obtained by measuring the RNA levels every 7 minutes for 119 minutes, thus a total of 18 time points, to cover two cycles. Since the dataset contained missing values, we complemented them by using the \texttt{imputeMissings} package in \texttt{R}. By complementing the dataset, we can use all $n = 1790$ genes and analyze the relationship between the RNA levels in the $q = 18$ time points and $p = 113$ TFs. We compared RVSManOpt with SeCURE(AL) and SeCURE(AE) by computing the number of selected experimentally confirmed TFs among the total number of the selected TFs and the proportion of experimentally confirmed TFs. It is known that there are 21 TFs which have been experimentally confirmed to be involved in the cell cycle regulation (Wang et al., 2007).

Table \ref{tb:yeast_cell_results} gives the results of a real data analysis. In RVSManOpt, the proportion of experimentally confirmed TFs is larger than both SeCURE(AL) and SeCURE(AE). RVSManOpt estimated $\hat{r} = 5$, while SeCURE(AL) and SeCURE(AE) estimated $\hat{r} = 4$. This result means that RVSManOpt may capture the latent structure of the yeast cell cycle data more precisely by identifying 5 latent factors.

Fig. \ref{fig:Yeast} shows estimated transcription levels of three of the experimentally confirmed TFs selected by RVSManOpt. The rest of the 12 experimentally confirmed TFs are available as the supplementary materials. Fig. \ref{fig:Yeast} indicates that the estimated transcription levels followed two cycles. It was experimentally confirmed that the transcription levels in the cell cycle did cover a two cycle time period. Thus, RVSManOpt was demonstrated to accurately estimate the cycles of data.

\begin{table}[H]
    \begin{center}
    \caption{Results of analysis of yeast cell cycle dataset.}
    \begin{tabular*}{\textwidth}{@{\extracolsep{\fill}} c|ccc}
        \hline
    Method &  \begin{tabular}{c}Total number of\\ selected TFs \end{tabular}& \begin{tabular}{c}Total number of\\ selected confirmed TFs\end{tabular} & \begin{tabular}{c}Proportion of\\ experimentally confirmed TFs\end{tabular}\\
        \hline
        \hline
        RVSManOpt & 15 & 67 & 0.224\\
        SeCURE(AL) & 17 & 83 &  0.205\\
        SeCURE(AE) & 17 & 83 & 0.205\\
        \hline
    \end{tabular*}
    \label{tb:yeast_cell_results}
    \end{center}
\end{table}

\begin{figure}[H]
    \centering
    \subcaptionbox{ACE2\label{fig:ACE2}}
    {\includegraphics[width=.32\linewidth]{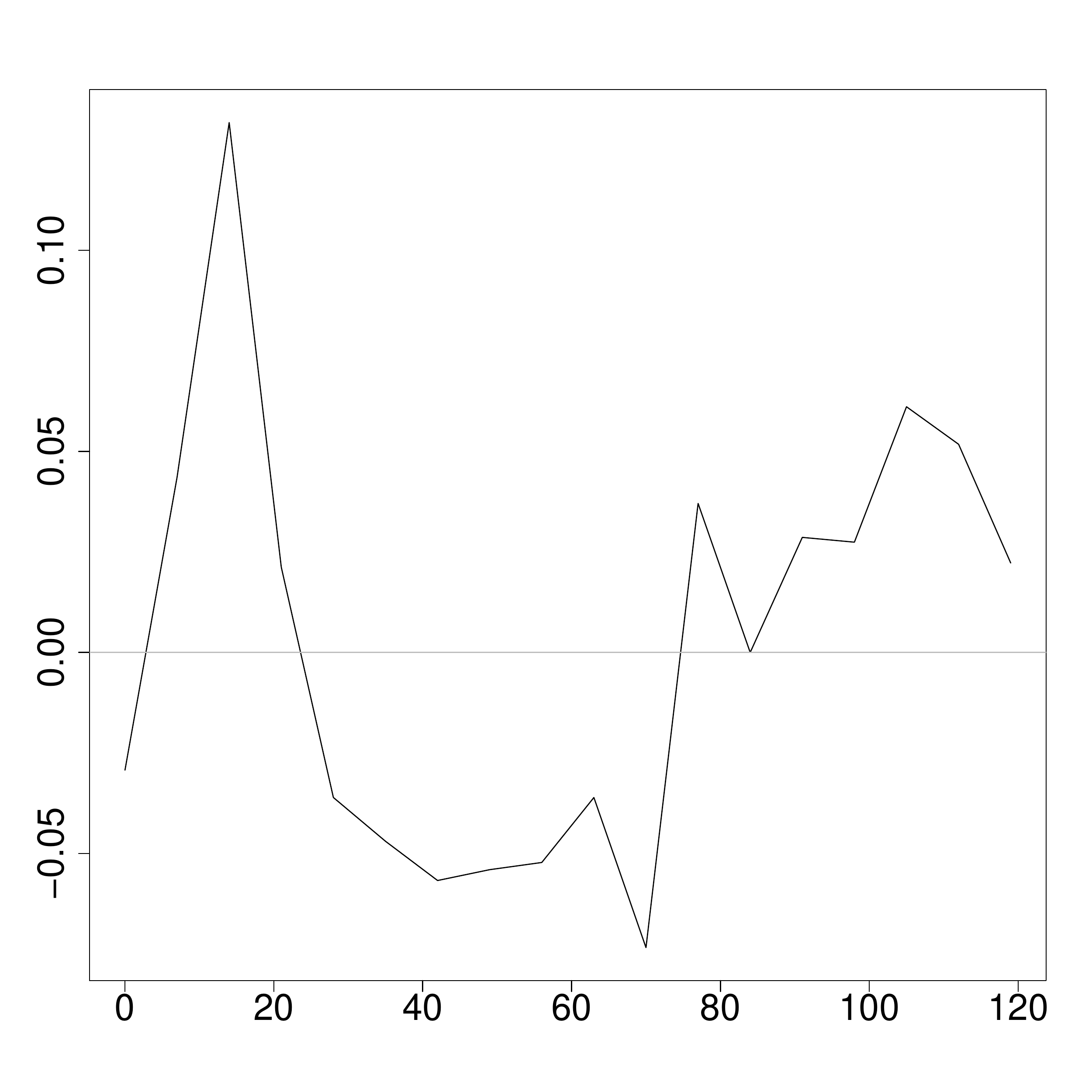}}
    \subcaptionbox{BAS1\label{fig:BAS1}}
    {\includegraphics[width=.32\linewidth]{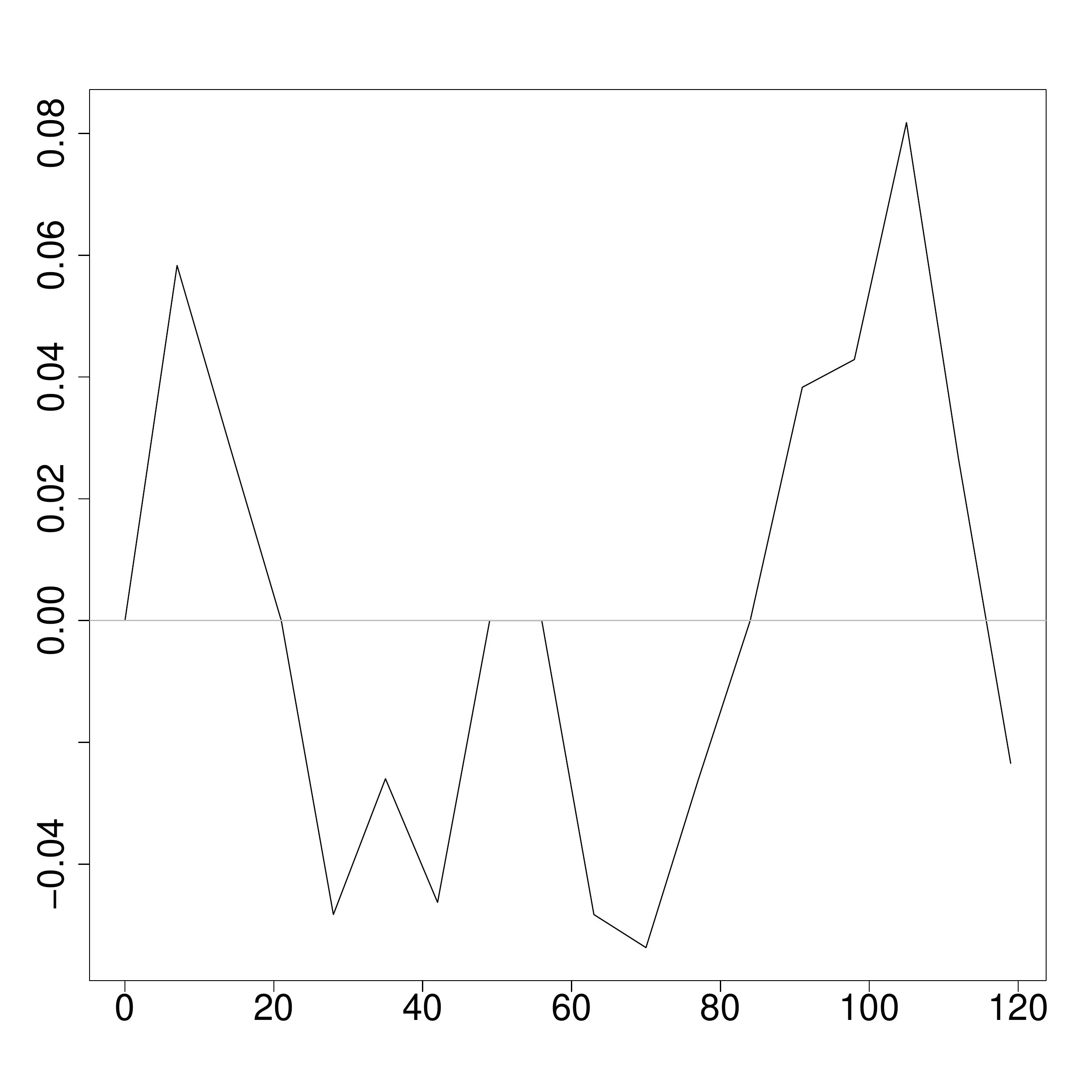}}
    \subcaptionbox{MBP1\label{fig:MBP1}}
    {\includegraphics[width=.32\linewidth]{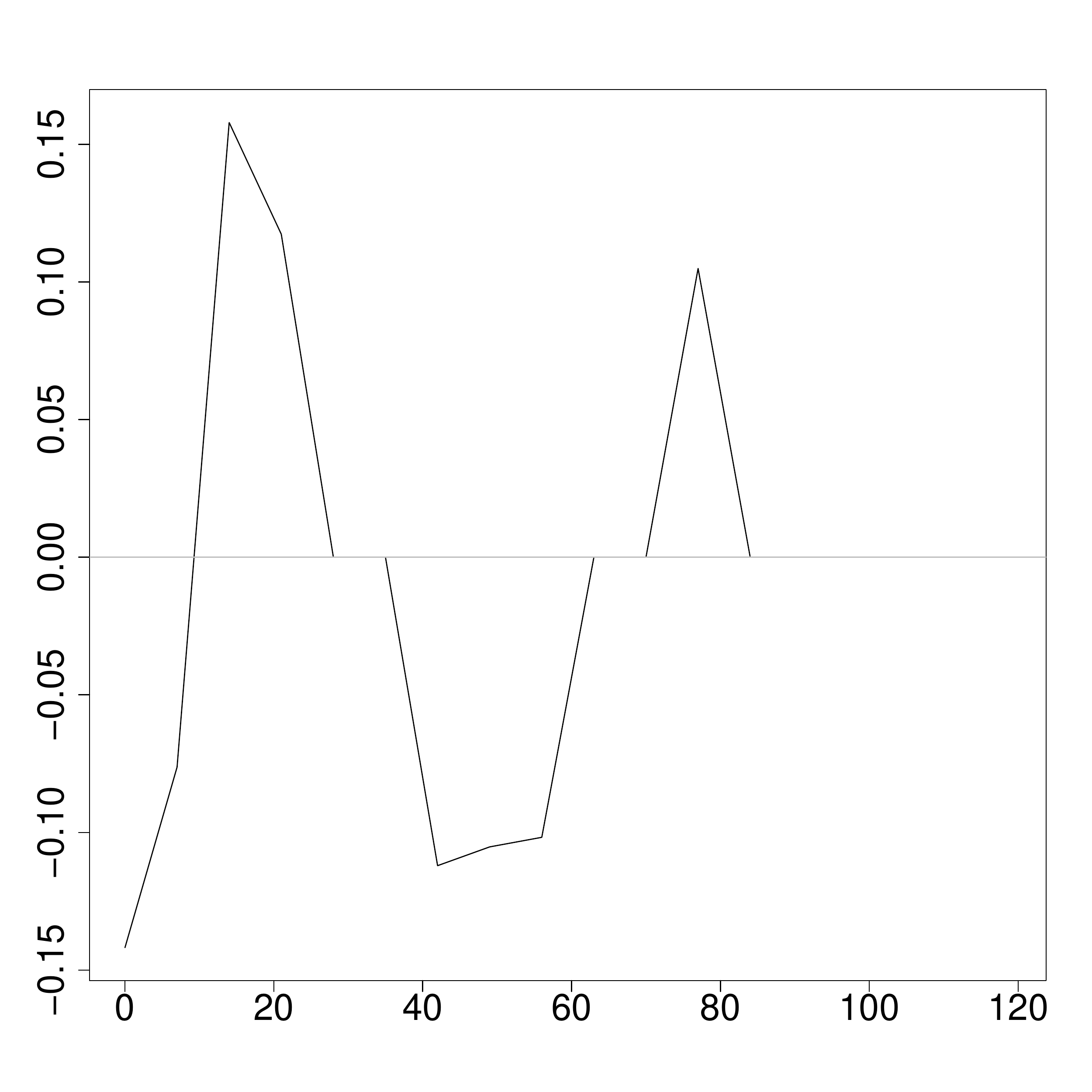}}
    \caption{Plots of estimated transcription levels of 3 experimentally confirmed TFs selected by RVSManOpt.}\label{fig:Yeast}
\end{figure}

\section{Concluding Remarks}
We proposed a minimization problem of SFAR on a Stiefel manifold and developed the factor extraction algorithm with rank and variable selection via sparse regularization and manifold optimization (RVSManOpt). RVSManOpt surpassed the traditional estimation procedure, which fails when the rank of the coefficient matrix is high. Numerical comparisons including Monte Carlo simulations and a real data analysis supported the usefulness of RVSManOpt.

In general, it is challenging to estimate parameters while preserving both orthogonality and sparsity. Mishra et al. (2017) indicates that enforcing orthogonality collapses sparsity and does not work from the viewpoint of prediction. Therefore, it may be unnecessary to construct a model with perfect orthogonality if we focus on prediction. Also, the recent paper by Absil and Hosseini (2019) discusses a theory of manifold optimization for non-smooth functions. It would be interesting to develop RVSManOpt based on this theory. We leave these as future topics.

\appendix
\section*{Appendix: Detailed description of update procedures for the parameters}\label{apd:calc}
\section*{Formulas for updating U and V} \label{apd:U_V}
The Euclidean gradient $\nabla \mathcal{L}_{\mat{U}}$ can be calculated as follows:
\begin{align*}
    \nabla \mathcal{L}_{\mat{U}} &= \frac{\partial}{\partial \mat{U}} \left[ \frac{1}{2} \norm{\mat{Y} - \mat{X} \mat{U} \mat{D} \trans{\mat{V}}}_F^2 + \frac{\gamma_1}{2} \norm{\mat{U} - \mat{U}^* + \mat{\Omega}}_F^2 \right] \nonumber \\
    &=  \frac{\partial}{\partial \mat{U}} \left[ \frac{1}{2} \Tr\left(  \mat{Y}\trans{\mat{Y}} -  2\mat{Y} \mat{V} \mat{D} \trans{\mat{U}} \trans{\mat{X}} + \mat{X} \mat{U} \mat{D}^2 \trans{\mat{U}} \trans{\mat{X}} \right)\right]  + \gamma_1 \left( \mat{U} - \mat{U}^* + \mat{\Omega} \right) \nonumber \\
    &= \frac{\partial}{\partial \mat{U}} \left[  \frac{1}{2}\Tr\left(\trans{\mat{U}} \trans{\mat{X}}\mat{X} \mat{U} \mat{D}^2  \right)  - \Tr \left( \mat{Y} \mat{V} \mat{D} \trans{\mat{U}} \trans{\mat{X}} \right) \right] + \gamma_1 \left( \mat{U} - \mat{U}^* + \mat{\Omega} \right)  \nonumber \\
    &= \frac{\partial}{\partial \mat{U}} \left[ \frac{1}{2}\Tr\left(n \mat{D}^2  \right)  - \Tr \left(  \trans{\mat{X}} \mat{Y} \mat{V} \mat{D} \trans{\mat{U}} \right) \right] + \gamma_1 \left( \mat{U} - \mat{U}^* + \mat{\Omega} \right) ,\quad \left(\because \trans{\mat{U}}\left(\frac{\trans{\mat{X}} \mat{X}}{n}\right) \mat{U} = \mat{I}_r  \right) \nonumber \\
    &= - \trans{\mat{X}} \mat{Y} \mat{V} \mat{D} + \gamma_1 \left( \mat{U} - \mat{U}^* + \mat{\Omega} \right).
\end{align*}
The formula for updating $\mat{U}$ is given by 
\begin{align*}
    \hat{\mat{U}} \gets \mathcal{R}_{\mat{U}}(-t_u \mathrm{grad} \mathcal{L}_{\mat{U}}),
\end{align*}
where $\mathcal{R}_{\mat{U}}$ is the retraction mapping on a generalized Stiefel manifold, $t_u$ is the Armijo step size, and $\mathrm{grad} \mathcal{L}_{\mat{U}}$ is the gradient on the generalized Stiefel manifold. $\mathrm{grad} \mathcal{L}_{\mat{U}}$ can be obtained by projecting the Euclidean gradient $\nabla \mathcal{L}_{\mat{U}}$ into the tangent space $\mathcal{T}_\mat{U} \StG{p}{r}$ by using projection operator $\mathcal{P}_\mat{U}(\cdot)$.

In a similar way, the Euclidean gradient $\nabla \mathcal{L}_{\mat{V}}$ can be calculated as follows:
\begin{align*}
    \nabla \mathcal{L}_{\mat{V}} &= \frac{\partial}{\partial \mat{V}} \left[ \frac{1}{2} \| \mat{Y} - \mat{X}\mat{U}\mat{D} \trans{\mat{V}} \|_F^2 + \frac{\gamma_2}{2} \norm{\mat{V} - \mat{V}^* + \mat{\Phi}}_F^2 + \frac{\gamma_3}{2} \norm{\mat{V} - \mat{V}^{**} + \mat{\Psi}}_F^2 \right] \nonumber \\
    &=  \frac{1}{2} \frac{\partial}{\partial \mat{V}} \left[\Tr\left(  \mat{Y}\trans{\mat{Y}} -  2\mat{Y} \mat{V} \mat{D} \trans{\mat{U}} \trans{\mat{X}} + \mat{X} \mat{U} \mat{D}^2 \trans{\mat{U}} \trans{\mat{X}} \right)\right] \nonumber\\
    & + \gamma_2 \left(\mat{V} - \mat{V}^* + \mat{\Phi} \right) + \gamma_3 \left(\mat{V} - \mat{V}^{**} + \mat{\Psi} \right)\nonumber \\
    &= -  \frac{\partial}{\partial \mat{V}} \left[   \Tr \left( \mat{Y} \mat{V} \mat{D} \trans{\mat{U}} \trans{\mat{X}} \right) \right] + \gamma_2 \left(\mat{V} - \mat{V}^* + \mat{\Phi} \right) + \gamma_3 \left(\mat{V} - \mat{V}^{**} + \mat{\Psi} \right) \nonumber \\
    &= - \frac{\partial}{\partial \mat{V}} \left[  \Tr \left(\mat{D} \trans{\mat{U}} \trans{\mat{X}} \mat{Y} \mat{V}  \right) \right] + \gamma_2 \left(\mat{V} - \mat{V}^* + \mat{\Phi} \right) + \gamma_3 \left(\mat{V} - \mat{V}^{**} + \mat{\Psi} \right) \nonumber \\
    &= - \trans{\mat{Y}} \mat{X} \mat{U} \mat{D} + \gamma_2 \left(\mat{V} - \mat{V}^* + \mat{\Phi} \right) + \gamma_3 \left(\mat{V} - \mat{V}^{**} + \mat{\Psi} \right).
\end{align*}
The formula for updating $\mat{V}$ is given by 
\begin{align*}
    \hat{\mat{V}} \gets \mathcal{R}_{\mat{V}}(-t_v \mathrm{grad} \mathcal{L}_{\mat{V}}),
\end{align*}
where $\mathcal{R}_{\mat{V}}$ is the retraction mapping on a Stiefel manifold, $t_v$ is the Armijo step size, and $\mathrm{grad} \mathcal{L}_{\mat{V}}$ is the gradient on the Stiefel manifold. $\mathrm{grad} \mathcal{L}_{\mat{V}}$ can be obtained by projecting the Euclidean gradient $\nabla \mathcal{L}_{\mat{V}}$ into the tangent space $\mathcal{T}_\mat{V} \St{q}{r}$ by using projection operator $\mathcal{P}_\mat{V}(\cdot)$. 

\section*{Formula for updating D}
The Euclidean gradient $\nabla \mathcal{L}_{\mat{D}}$ is given by
    \begin{align*}
        \nabla \mathcal{L}_{\mat{D}} &= \frac{\partial}{\partial \mat{D}} \left[ \frac{1}{2} \norm{\mat{Y} - \mat{X} \mat{U} \mat{D} \trans{\mat{V}}}_F^2 \right]  \nonumber \\
       &=\frac{1}{2} \frac{\partial}{\partial \mat{D}}\left[ \Tr\left(  \mat{Y}\trans{\mat{Y}} -  2\mat{Y} \mat{V} \mat{D} \trans{\mat{U}} \trans{\mat{X}} + \mat{X} \mat{U} \mat{D}^2 \trans{\mat{U}} \trans{\mat{X}} \right)\right] \nonumber \\
       &= \frac{1}{2} \frac{\partial}{\partial \mat{D}}\left[ \Tr\left(\mat{X} \mat{U} \mat{D}^2 \trans{\mat{U}} \trans{\mat{X}}  -  2\mat{Y} \mat{V} \mat{D} \trans{\mat{U}} \trans{\mat{X}} \right) \right] \nonumber \\
       &= \frac{1}{2} \frac{\partial}{\partial \mat{D}}\left[ \Tr\left( \trans{\mat{U}}\trans{\mat{X}} \mat{X} \mat{U} \mat{D}^2  -  2\trans{\mat{U}} \trans{\mat{X}} \mat{Y} \mat{V} \mat{D}  \right)\right] \nonumber \\
       &= \frac{1}{2} \frac{\partial}{\partial \mat{D}} \left[ \Tr\left( n \mat{D}^2  -  2\trans{\mat{U}} \trans{\mat{X}} \mat{Y} \mat{V} \mat{D}  \right) \right] ,\quad \left(\because \trans{\mat{U}}\left(\frac{\trans{\mat{X}} \mat{X}}{n}\right) \mat{U} = \mat{I}_r  \right) \nonumber \\
       &= n\mat{D} - \trans{\mat{V}} \trans{\mat{Y}} \mat{X} \mat{U}.
    \end{align*}
    When $\nabla \mathcal{L}_{\mat{D}}= \mat{0}$, the optimal solution of $\hat{\mat{D}}$ is given by
    \begin{align*}
        n \hat{\mat{D}} - \trans{\mat{V}} \trans{\mat{Y}} \mat{X} \mat{U} = \mat{0} \nonumber \\
        \hat{\mat{D}} =  \frac{1}{n} \trans{\mat{V}} \trans{\mat{Y}} \mat{X} \mat{U}.
    \end{align*}
    Therefore, the formula for updating $\mat{D}$ is given by
    \begin{align*}
        \hat{\mat{D}} \gets  \mathrm{diag} \left(\frac{1}{n} \trans{\mat{V}} \trans{\mat{Y}} \mat{X} \mat{U} \right).
    \end{align*}

\section*{Formulas for updating $\mat{U}^*$ and $\mat{V}^*$}
The augmented Lagrangian with respect to $\mat{U}^*$ is given by
\begin{align*}
    \mathcal{L}(\mat{U}^*) = n\lambda_1 \sum_{i=1}^{p} \sum_{j=1}^q w^{(u)}_{ij}|u_{ij}^*| + \frac{\gamma_1}{2} \norm{\mat{U} - \mat{U}^* + \mat{\Omega}}_F^2.
\end{align*}
The partial derivative of $\mathcal{L}(\mat{U}^*)$ is calculated as follows:
    \begin{align*}
        \frac{\partial \mathcal{L}(\mat{U}^*)}{u_{ij}^*} = n\lambda_1 w_{ij}^{(u)} \partial |u_{ij}^*| + \gamma_1 (u_{ij} - u_{ij}^* + \omega_{ij}),
    \end{align*}
where $\partial | \cdot |$ is the subderivative operator defined as
    \begin{align*}
        \partial | a | = \begin{cases}
            \{ -1 \}, & (a < 0),\\
            [-1, 1], & (a = 0),\\
            \{ 1 \}, & (a > 0).
        \end{cases}
    \end{align*}
    When this partial derivative is equal to $0$, the element of $\mat{U}^*$ is represented as
    \begin{align*}
        u_{ij}^* = u_{ij} + \omega_{ij} - \frac{n\lambda_1 w_{ij}^{(u)}}{\gamma_1} \partial |u_{ij}^*|.
    \end{align*}
    Thus, the formula for updating $\mat{U}^*$ can be obtained as follows:
    \begin{align*}
        u_{ij}^* = \begin{cases}
            u_{ij} + \omega_{ij} - \frac{n\lambda_1 w_{ij}^{(u)}}{\gamma_1}, & \left(u_{ij} + \omega_{ij} > \frac{n\lambda_1 w_{ij}^{(u)}}{\gamma_1} \right),\\
            0, & \left( |u_{ij} + \omega_{ij}| \leq \frac{n\lambda_1 w_{ij}^{(u)}}{\gamma_1} \right),\\
            u_{ij} + \omega_{ij} + \frac{n\lambda_1 w_{ij}^{(u)}}{\gamma_1}, & \left(u_{ij} + \omega_{ij} < \frac{n\lambda_1 w_{ij}^{(u)}}{\gamma_1} \right).
        \end{cases}
    \end{align*}
    This formula can be simplified using the soft-thresholding operator $\mathrm{S}(\cdot, \cdot)$ as follows:
\begin{align*}
    {u_{ij}^*} \gets \mathrm{S} \left(u_{ij} + \omega_{ij}, \frac{n\lambda_1 w_{ij}^{(u)}}{\gamma_1} \right),
\end{align*}
where $\omega_{ij}$ is the $(i,j)$-th element of $\Omega$ and $\mathrm{S}(\cdot, \cdot)$ is the soft-thresholding operator
    \begin{align*}
        \mathrm{S} \left( x, \lambda \right) &= \sign(x) (|x| - \lambda)_+,\quad (x)_+ = \max\left\{x, 0 \right\},
    \end{align*}
    \begin{align*}
        \sign(x) &= \begin{cases}
            1, & (x > 0),\\
            0, & (x = 0),\\
            -1, & (x < 0).
        \end{cases}
    \end{align*}

    In a similar way to the updating of $\mat{U}^*$, the formula for updating $\mat{V}^*$ can be obtained as follows:
    \begin{align*}
        v_{ij}^* = \mathrm{S} \left(v_{ij} + \phi_{ij}, \frac{n\alpha \lambda_2 w_{ij}^{(v)}}{\gamma_2} \right).
    \end{align*}
    where $\phi_{ij}$ is the $(i,j)$-th element of $\Phi$.

\section*{Formula for updating $\mat{V}^{**}$}
The augmented Lagrangian with respect to $\mat{V}^{**}$ is given by
\begin{align*}
    \mathcal{L}(\mat{V}^{**}) = n\sqrt{q} (1-\alpha) \lambda_2 \sum_{i=1}^{r} w_i^{(d)} \mathds{1}\left( \vec{v}_i^{**} \neq \vec{0} \right) + \frac{\gamma_3}{2} \norm{\mat{V} - \mat{V}^{**} + \mat{\Psi}}_F^2.
\end{align*}
Here, we consider the augmented Lagrangian for every column vector $\vec{v}_{i}^{**}$, $i=1,\dots,r$, as follows:
    \begin{align*}\label{eq:Lagrangian_rspct_vast2}
        \mathcal{L}(\vec{v}_i^{**}) = n\sqrt{q} (1-\alpha) \lambda_2 w_i^{(d)} \mathds{1}\left( \vec{v}_i^{**} \neq \vec{0} \right) + \frac{\gamma_3}{2} \norm{\vec{v}_i - \vec{v}_i^{**} + \vec{\psi}_i}_2^2.
    \end{align*}
This equation can be divided into $\vec{v}_i^{**} = \vec{0}$ and $\vec{v}_i^{**} \neq \vec{0}$ cases as follows:
    \begin{align*}
        \mathcal{L}(\vec{v}_i^{**}) = \begin{cases}
            \frac{\gamma_3}{2} \norm{\vec{v}_i + \vec{\psi}_i}_2^2, & \left( \vec{v}_i^{**} = \vec{0} \right),\\
            n\sqrt{q} (1-\alpha) \lambda_2 w_i^{(d)} + \frac{\gamma_3}{2} \norm{\vec{v}_i - \vec{v}_i^{**} + \vec{\psi}_i}_2^2, & \left( \vec{v}_i^{**} \neq \vec{0} \right).
        \end{cases}
    \end{align*}
    When $\vec{v}_i^{**} \neq \vec{0}$, the optimal solution $\vec{v}_i^{**}$ can be obtained as follows:
    \begin{align*}
        \vec{v}_i^{**} = \vec{v}_i + \vec{\psi}_i.
    \end{align*}
    When we substitute $\vec{v}_i + \vec{\psi}_i$ in for $\vec{v}_i^{**}$ in $\mathcal{L}(\vec{v}_i^{**})$, the value of $\mathcal{L}(\vec{v}_i^{**})$ is $\sqrt{q} (1-\alpha) \lambda_2 w_i^{(d)}$. It is necessary to satisfy the following condition
    \begin{align*}
        \frac{\gamma_3}{2} \norm{\vec{v}_i + \vec{\psi}_i}_2^2 \geq n\sqrt{q} (1-\alpha) \lambda_2 w_i^{(d)}.
    \end{align*}
   The formula for updating $\mat{v}_i^*$ can be obtained as follows:
    \begin{align*}
        \vec{v}_i^{**} = \begin{cases}
            \vec{0}, & \left( \norm{\vec{v}_i + \vec{\psi}_i}_2 \leq \sqrt{\frac{2n\sqrt{q}(1-\alpha)\lambda_2 w_i^{(d)})}{\gamma_3}} \right),\\
            \vec{v}_i + \vec{\psi}_i, & \left( \norm{\vec{v}_i + \vec{\psi}_i}_2 > \sqrt{\frac{2n\sqrt{q}(1-\alpha)\lambda_2 w_i^{(d)})}{\gamma_3}} \right).
        \end{cases}
    \end{align*}
    This formula is simplified by using the hard-thresholding operator $\mathrm{H}(\cdot, \cdot)$  as follows:
\begin{align*}
    {\vec{v}^{**}_{i}} \gets \mathrm{H} \left({\vec{v}}_{i} +  \vec{\psi}_{i}, \sqrt{\frac{2n\sqrt{q}(1-\alpha)\lambda_2 w_i^{(d)})}{\gamma_3}} \right),
\end{align*}
where $\psi_{ij}$ is the $(i,j)$-th element of $\Psi$ and $\mathrm{H}(\cdot, \cdot)$ is the hard-thresholding operator 
\begin{align*}
    \mathrm{H} \left( \vec{x}, \lambda \right) &= \mathds{1} \left(1 - \frac{\lambda}{\norm{\vec{x}}_2} > 0\right) \vec{x}.
\end{align*}

    \section*{Acknowledgments}
    S. K. was supported by JSPS KAKENHI Grant Number JP19K11854 and MEXT KAKENHI Grant Numbers JP16H06429, JP16K21723, and JP16H06430. The super-computing resource was provided by Human Genome Center (the Univ. of Tokyo).

\end{document}